\documentclass[]{spie}  

\usepackage{amsmath,amsfonts,amssymb, bm}
\usepackage{graphicx}
\usepackage[colorlinks=true, allcolors=blue]{hyperref}

\usepackage{lipsum}
\usepackage{color}
\usepackage[caption=false]{subfig} 

\usepackage{booktabs}   

\title{Mechanical design and fabrication of a kinetic sculpture with application to bioinspired drone design}
\author{Andrew Lessieur}
\author{Eric Sihite}
\author{Pravin Dangol}
\author{Akshath Singhal}
\author{Alireza Ramezani}
\affil{SiliconSynapse Laboratory at Northeastern University, Boston, USA}

\authorinfo{Further author information: (Send correspondence to Alireza Ramezani)\\  Alireza Ramezani: E-mail: a.ramezani@northeastern.edu, Telephone: 1 (734) 604 1214}

\begin{document} 
\maketitle

\begin{abstract}
Biologically-inspired robots are a very interesting and difficult branch of robotics dues to its very rich dynamical and morphological complexities. Among them, flying animals, such as bats, have been among the most difficult to take inspiration from as they exhibit complex wing articulation. We attempt to capture several of the key degrees-of-freedom that are present in the natural flapping gait of a bat. In this work, we present the mechanical design and analysis of our flapping wing robot, the Aerobat, where we capture the plunging and flexion-extension in the bat's flapping modes. This robot utilizes gears, cranks, and four-bar linkage mechanisms to actuate the arm-wing structure composed of rigid and flexible components monolithically fabricated using PolyJet 3D printing. The resulting robot exhibits wing expansion and retraction during the downstroke and upstroke respectively which minimizes the negative lift and results in a more efficient flapping gait.
\end{abstract}


\section{INTRODUCTION}
\label{sec:intro}  


The overall goal of this work is to advance the theory and practice of aerial robots that are soft, agile, collision-tolerant, and energetically efficient by the biomimicry of key airborne vertebrate flight characteristics. In recent years, much attention has been drawn to make our residential and work spaces smarter and to materialize the concept of smart cities \cite{everaerts2008use}. As a result, safety and security aspects are gaining ever growing importance \cite{pavlidis2001urban} and drive a lucrative market. Systems that can provide situational awareness to humans in residential and work spaces or contribute to dynamic traffic control in cities will result in large-scale intelligent systems with enormous societal impact and economic benefit. 

Current state-of-the-art solutions with rotary or fixed-wing features fall short in addressing the challenges and pose extreme dangers to humans. Fixed or rotary-wing systems are widely applied for surveillance and reconnaissance, and there is a growing interest to add suites of on-board sensors to these unmanned aerial systems (UAS) and use their aerial mobility to monitor and detect hazardous situations in residential spaces. While, these systems, e.g., quadrotors, can demonstrate agile maneuvers and have demonstrated impressive fault-tolerance in aggressive environments, quadrotors and other rotorcrafts require a safe and collision-free task space for operation since they are not collision-tolerant due to their rigid body structures. The incorporation of soft and flexible materials into the design of such systems has become common in recent years, yet, the demands for aerodynamic efficiency prohibit the use of rotor blades or propellers made of extremely flexible materials. 

The flight apparatus of birds and bats can offer invaluable insights into novel micro aerial vehicle (MAV) designs that can safely operate within residential spaces. The pronounced body articulation (morphing ability) of these flyers is key to their unparalleled capabilities. These animals can reduce the wing area during upstrokes and can extend it during downstrokes to maximize positive lift generation \cite{tobalske2000biomechanics}. It is known that some species of bats can use differential inertial forces to perform agile zero-angular momentum turns \cite{riskin2012upstroke}. Biological studies suggest that the articulated musculoskeletal system of animals can absorb impact forces therefore can enhance their survivability in the event of a collision \cite{roberts2013tendons}.

The objective of this paper is to design and develop a bio-inspired soft and articulated armwing structure which will be an integral component of a morphing aerial co-bot. In our design, we draw inspiration from bats. Bat membranous wings possess unique functions \cite{tanaka2015flexible} that make them a good example to take inspiration from and transform current aerial drones. In contrast with other flying vertebrates, bats have an extremely articulated musculoskeletal system, key to their body impact-survivability and deliver an impressively adaptive and multimodal locomotion behavior \cite{riskin2008quantifying}. Bats exclusively use this capability with structural flexibility to generate the controlled force distribution on each wing membrane. The wing flexibility, complex wing kinematics, and fast muscle actuation allow these creatures to change the body configuration within a few tens of milliseconds. These characteristics are crucial to the unrivaled agility of bats \cite{azuma2006biokinetics} and copying them can potentially transform the state-of-the-art aerial drone design. 

As part of our past work, we have developed a flapping wing robot mimicking bat flight, the \textit{BatBot} (B2) \cite{ramezani_biomimetic_2017, ramezani_lagrangian_2015, hoff_synergistic_2016,ramezani_nonlinear_2016, hoff_reducing_2017, ramezani_describing_2017, syed_rousettus_2017, hoff_optimizing_2018,  hoff_trajectory_2019, ramezani_towards_2020}, which incorporates a linkage mechanism to articulate plunging motion and wing folding through mediolateral movements. In our most recent work, we have developed a computational structure fabricated monolithically using soft and rigid components \cite{sihite_computational_2020, sihite_enforcing_2020}, as shown in Fig.~\ref{fig:cover_photo}. This structure, called the \textit{Kinetic Sculpture} (KS), was designed to capture bat dynamically versatile wing conformations and mimic the soft and articulated armwing structure of bat membraneous wing through \textit{mechanical intelligence}.

\section{ARMWING STRUCTURE DESIGN OVERVIEW}

\begin{figure}[!t]
    \centering
    \includegraphics[width=\linewidth]{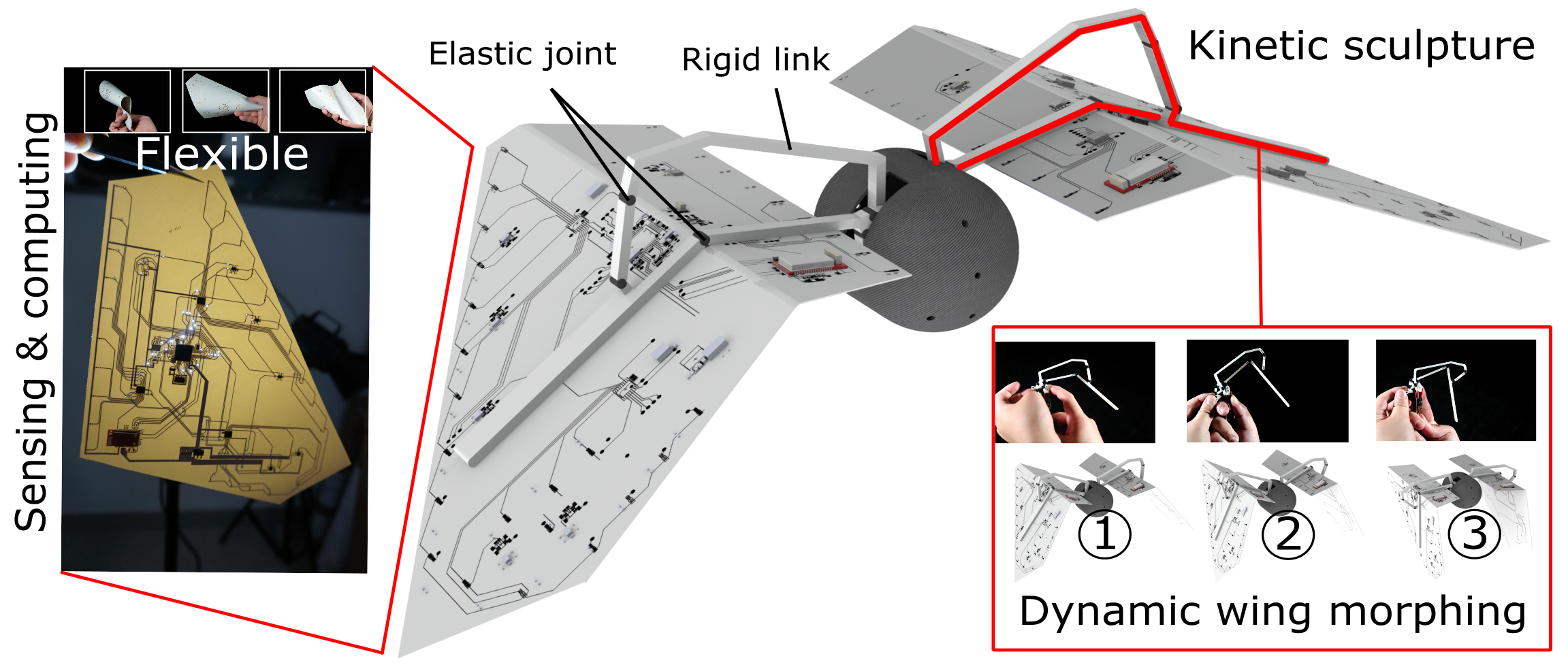}
    \caption{Illustration of Northeastern University's \textit{Aerobat}, which features a kinetic sculpture capturing the dynamic wing morphing of bats. The wing membrane is made out of flexible PCB which carries sensors, motor drivers, communication unit, and a microcontroller.}
    \label{fig:cover_photo}
\end{figure}

The Northeastern University's \textit{Aerobat}, shown in Fig.~\ref{fig:cover_photo}, is a flapping wing robot with a maximum wingspan of 30 cm. This robot is designed for evaluating the flapping performance of the KS shown in Fig. \ref{fig:cover_photo}. This design shows that we are capable of animating dynamic wing morphing seen in bats. The KS design iterations and robot design will be discussed in more detail in this section.

\begin{figure}[t]
    \centering
    \includegraphics[width=0.6\linewidth]{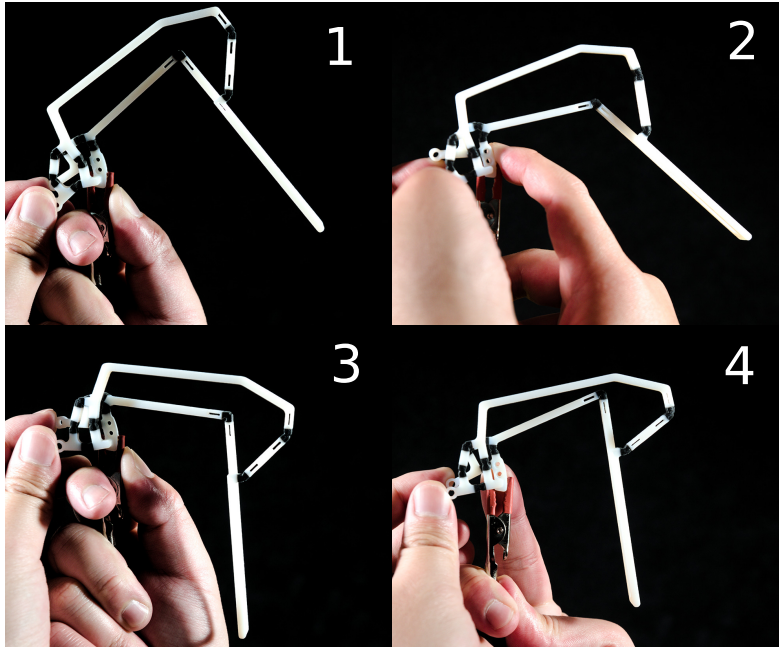}
    \caption{The time-lapsed images of the bat armwing articulation. Images 1-2 show the wing downstroke/expansion motion while images 3-4 show the wing upstroke/retraction motion. The armwing is monolithically fabricated using both rigid and flexible materials in a PolyJet 3D printer.}
    \vspace{-0.1in}
    \label{fig:wing_timelapse}
\end{figure}

The Aerobat is designed to show the feasibility of untethered flight using the KS we designed in Section \ref{sec:ks}. In this prototype, the KS is implemented using using rigid linkages and joints fabricated out of acrylic lasercut components and metal hinges. This linkage structure is capable of articulating our desired wing folding and expansion, as shown in \mbox{Fig.~\ref{fig:wing_timelapse}}. The body and several other mounting components, such as spacers, were fabricated out of 3D printed plastic.

We designed the wing membrane using flexible PCB, shown in Fig. \ref{fig:cover_photo}, where we attach sensors, motor drivers, communication modules, and microcontroller unit. The components attached on the wings will be used as a part of our future work to incorporate mechanical intelligence to facilitate control in this robot. In this framework, we utilize several small and low power actuators, called the Feedback Driven Components (FDC), to adjust the KS conformations. By attaching these FDCs on a highly sensitive KS component, we can significantly change the resulting flapping trajectory through a small adjustment in KS conformation. This allow us to adjust the resulting aerodynamic forces and facilitate control to stabilize the robot's attitude. our robot is pitch unstable which is what we expected in a tail-less ornithopthers, which makes active or passive stabilizer necessary in this robot.

The KS of both wings are coupled using 7 spur gears, allowing us to actuate them using a single electro-magnetic actuator. We selected the fastest and lightest brushless motor with a gearbox reduction ratio of 75:1 that is capable of providing enough speed and torque to fully actuate the KS at high flapping speed. 

\subsection{Kinetic Sculpture Design}
\label{sec:ks}

We designed our KS using flexible and rigid materials monolithically fabricated using PolyJet 3D printing \cite{sihite_computational_2020}. The KS utilized a series of four-bar mechanisms (mechanical amplifiers) to animate the wing plunging motion in addition to the wing folding and expansion. We designed the driving mechanism such that both wings can be actuated by a single motor through the use of spur gears and crank mechanisms.

We used the PolyJet flexible materials for the monolithic fabrication process. These materials come in Shore hardness scale range of 30A to 85A which is formed by mixing the rigid \textit{Vero White} and the flexible \textit{Agilus Black}. The design variations for compliant joints outlined in [\citenum{trease2005design}] were considered where they vary in size, off-axis stiffness, axis drift, stress concentration, and range of motion. 

We picked a combination of 1.3 and 2mm thick flexible hinges and the flexible materials \cite{polyjet_datasheet} shown in Table~\ref{tab:hinge_material}. The larger cross-sectional areas improve the durability and the off-axis stiffness of the joints, but also increased the overall weight of the KS. We found out that the KS with durometer scale of 50A is too soft to be used in the wing articulation and it has very little off-plane and torsional stiffness even in the thicker hinge design. On the other hand, the KS with durometer scale of 85A and 2mm hinge thickness is too stiff and brittle. Out of all the combinations we tested, we found that the KS with 70A durometer and 1.3 mm hinge thickness have the best overall result in the wing articulation and stiffness.

\begin{table}[t]
    \centering
    \caption{Table of the Tested Hinge Materials}
    \vspace{0.1in}
    \begin{tabular}{ c | c c}
    \toprule
    Material & Shore Hardness (A) & Elongation at Break (\%) \\
    \midrule
    FLX 9850 & 50 - 55 & 170 - 210 \\
    FLX 9870 & 60 - 70 & 120 - 140 \\
    FLX 9885 & 80 - 85 & 70 - 90 \\
    \bottomrule
    \end{tabular}
    \label{tab:hinge_material}
\end{table}

\subsection{Armwing Optimization}

\begin{figure}[t]
    \centering
    \subfloat[Unoptimized armwing angles.]{%
    \label{fig:plot_wing_angles_a}\includegraphics[clip,width=0.45\columnwidth]{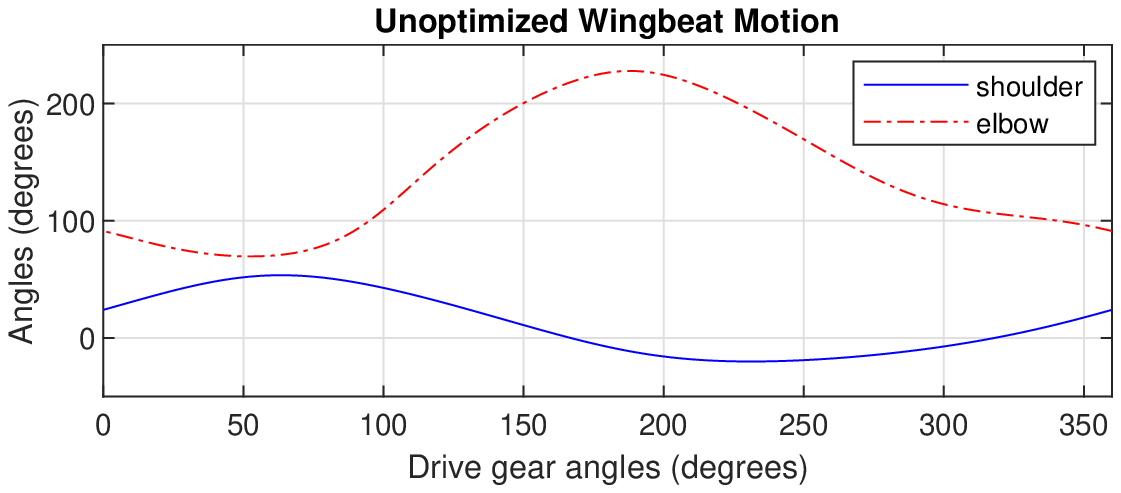}%
    }
    \subfloat[Optimized armwing angles vs. target trajectories.]{%
    \label{fig:plot_wing_angles_b}\includegraphics[clip,width=0.45\columnwidth]{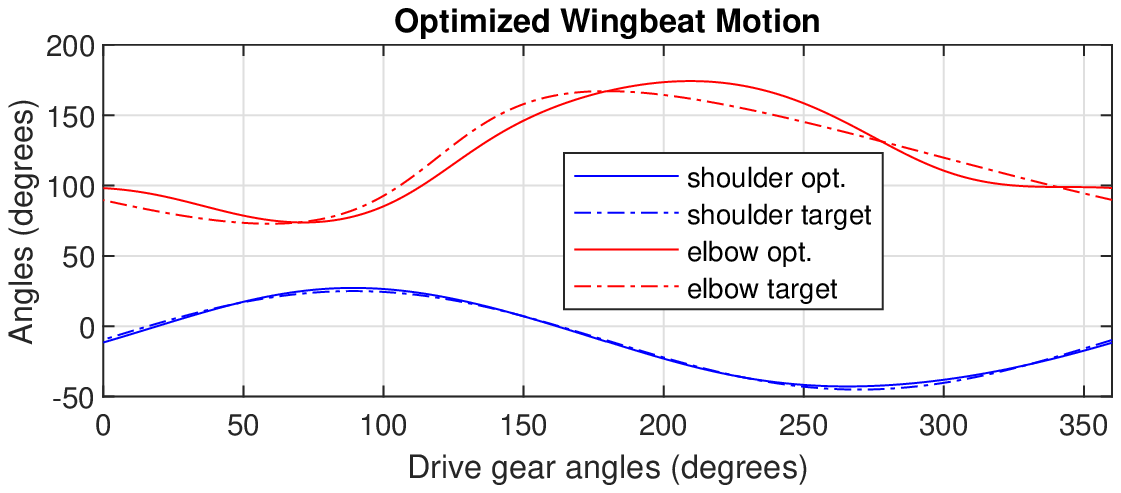}%
    }
    
\caption{The unoptimized and optimized armwing shoulder and elbow angles within a single wingbeat. The optimization has successfully found the parameters that results in close tracking to the target trajectory.}
\label{fig:plot_wing_angles}
\end{figure}

The ideal flapping motion that we are looking for has the following properties: (1) the wing extends and retracts during downstroke and upstroke respectively, (2) the wing is already partially expanded before the downstroke motion begins. The desired shoulder and elbow trajectories ($\hat{\theta}_s$ and $\hat{\theta}_e$, respectively) can be seen in Fig. \ref{fig:plot_wing_angles_b}, which are defined as the following sinusoidal functions
\begin{equation}
\begin{gathered}
\hat \theta_s = 35^\circ\, \sin(\phi) - 10^\circ\\ 
\hat \theta_e = -0.5\,\tan^{-1}\left( \tfrac{-0.5\,\sin(\phi + 2\pi/3)}{1 + 0.5\,\cos(\phi+2\pi/3)}  \right) \, 45^\circ + 120^\circ,
\end{gathered}
\end{equation}
where $\phi \in [0,2\pi)$. $\hat{\theta}_e$ is a skewed sinusoidal function which allows the wing to expands faster than the retraction and have a full wingspan in the middle of the downstroke.

The design optimization will solve for some of the mechanism design parameters using our initial mechanism design in Solidworks for the initial conformation parameter $\bm{q}$. There are 38 parameters in the design space of this armwing and we constrain some of these parameters to fit our design criterion and reduce the search space of the optimizer. In order to have a symmetric gait between the left and right wing, the drive gears must be centered and the crank arm maximum horizontal length must be aligned with the body $y$ axis. Additionally, we fix the values for the gear positions and the length of humerus and radius linkages. This leaves us with 30 design parameters to optimize. 

Considering the large design space of this wing structure, solving for all 30 parameters at the same time is not practical due to the large computational time and search space. The radius mechanism must follow a trajectory in relation to the humerus mechanism to articulate the appropriate elbow angle. Therefore, we can separately optimize the humerus and radius mechanisms, starting from the humerus mechanism. The humerus and radius mechanisms have 13 and 17 design parameters, respectively.

The optimization problem can be formulated as
\begin{equation}
\begin{aligned}
\min_{\bm{q}} \quad & (\bm{y}^T\,\bm{y}) / N \\
\textrm{subject to:} \quad &\bm{q}_{min} \leq \bm{q} \leq \bm{q}_{max}, \quad \bm{f}_c \leq 0,
\end{aligned}
\label{eq:op_problem}
\end{equation}
where the the cost function is the mean squared value of $\bm{y}$ which is the difference between target vs. the simulated trajectory, $N$ is the data size, $\bm{q}$ is the parameter to optimize, $\bm{q}_{min}$ and $\bm{q}_{max}$ are the parameter bounds, and $\bm{f}_c$ is the constraint function. We used the interior-point method as the optimization algorithm in Matlab which has successfully found a solution that matches the target trajectory well, as shown in Fig.~\ref{fig:plot_wing_angles}. 

\subsection{Sensitivity Analysis}

\begin{figure*}[t]
\centering
    \subfloat[Adjustment to $l_1$ length.]{%
    \label{fig:plot_sensitivity_1}\includegraphics[clip,width=0.32\linewidth]{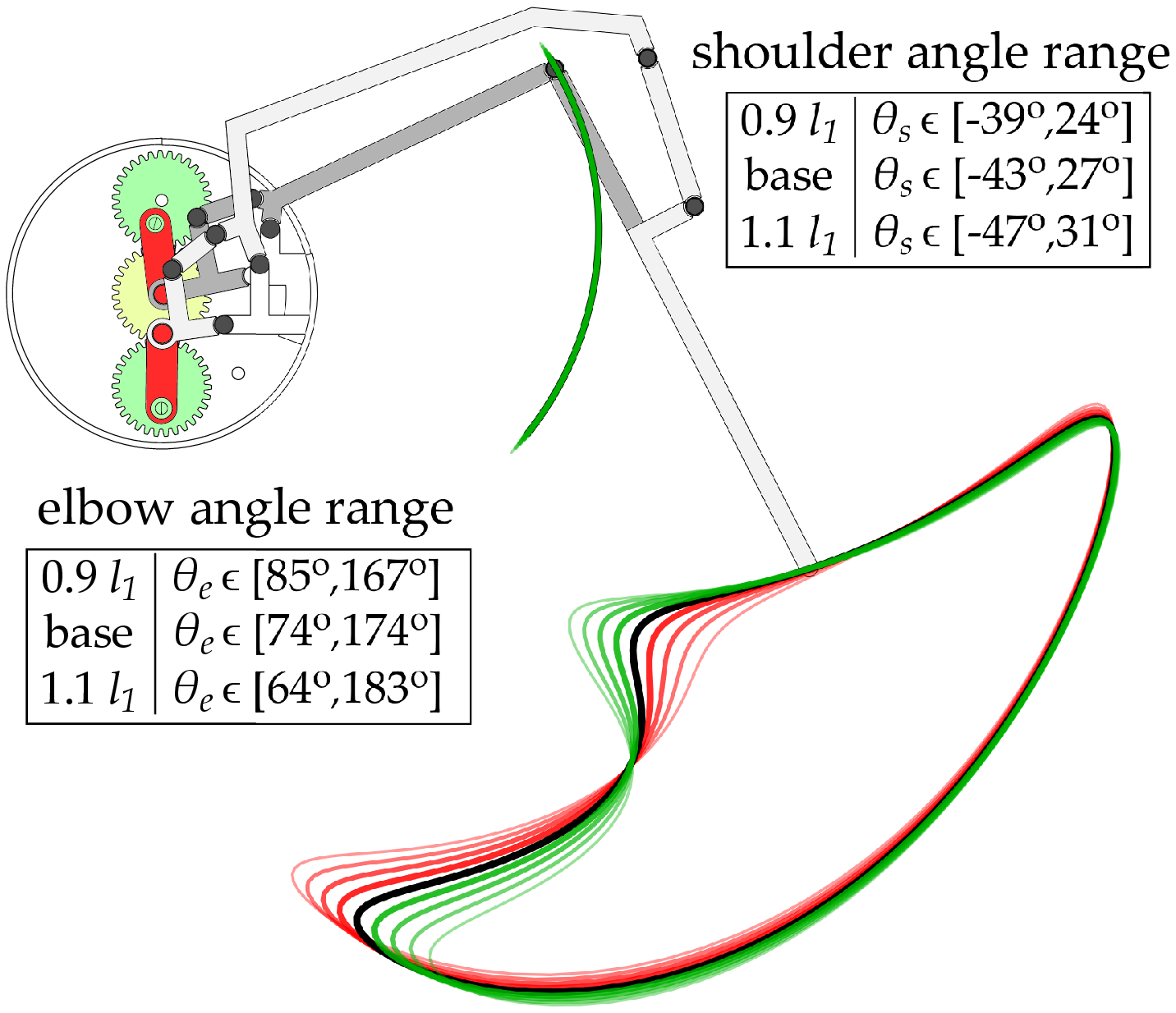}%
    } \hfil
    \subfloat[Adjustment to $l_4$ length.]{%
    \label{fig:plot_sensitivity_2}\includegraphics[clip,width=0.32\linewidth]{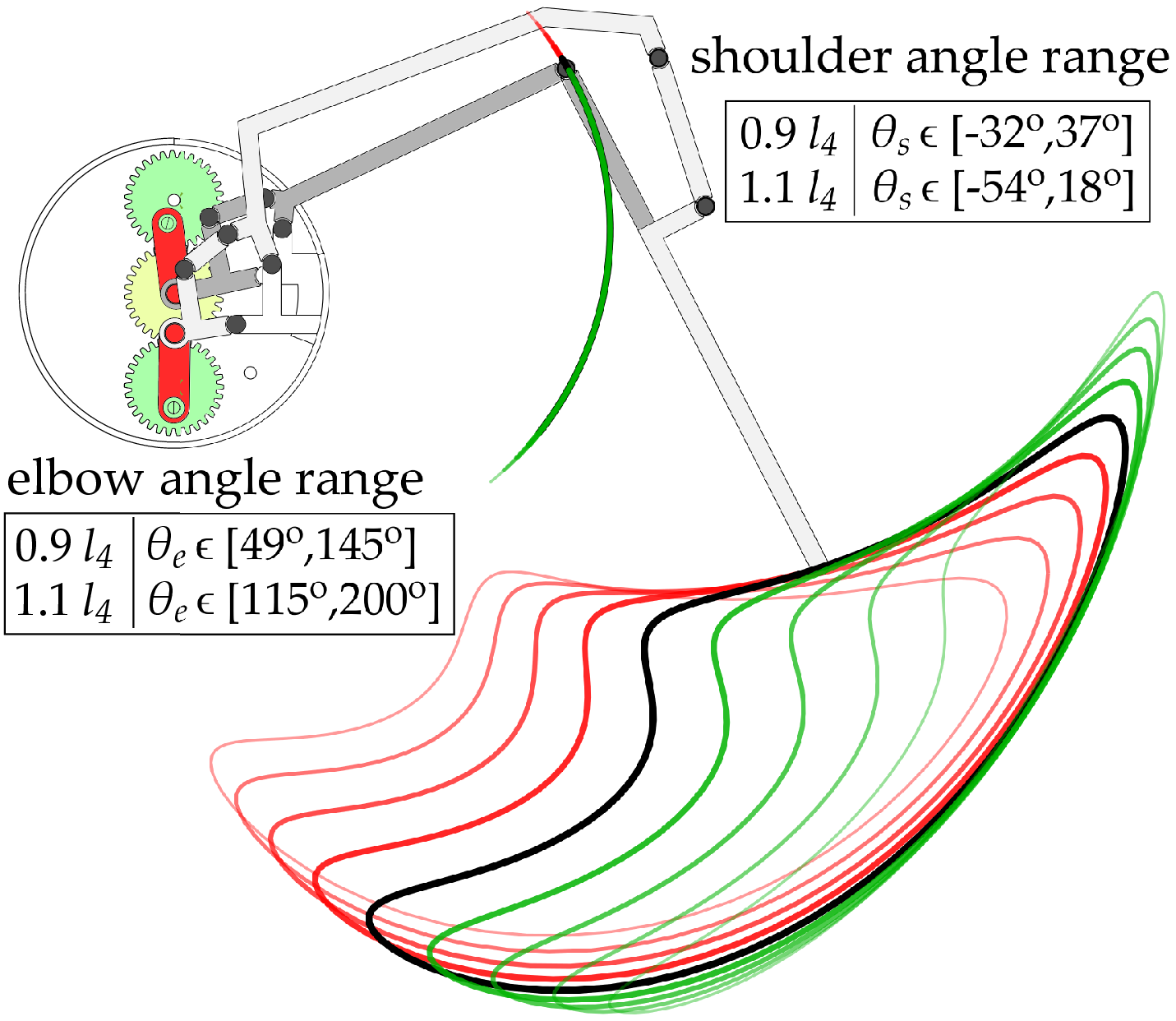}%
    } \hfil
    \subfloat[Adjustment to $l_9$ length.]{%
    \label{fig:plot_sensitivity_3}\includegraphics[clip,width=0.32\linewidth]{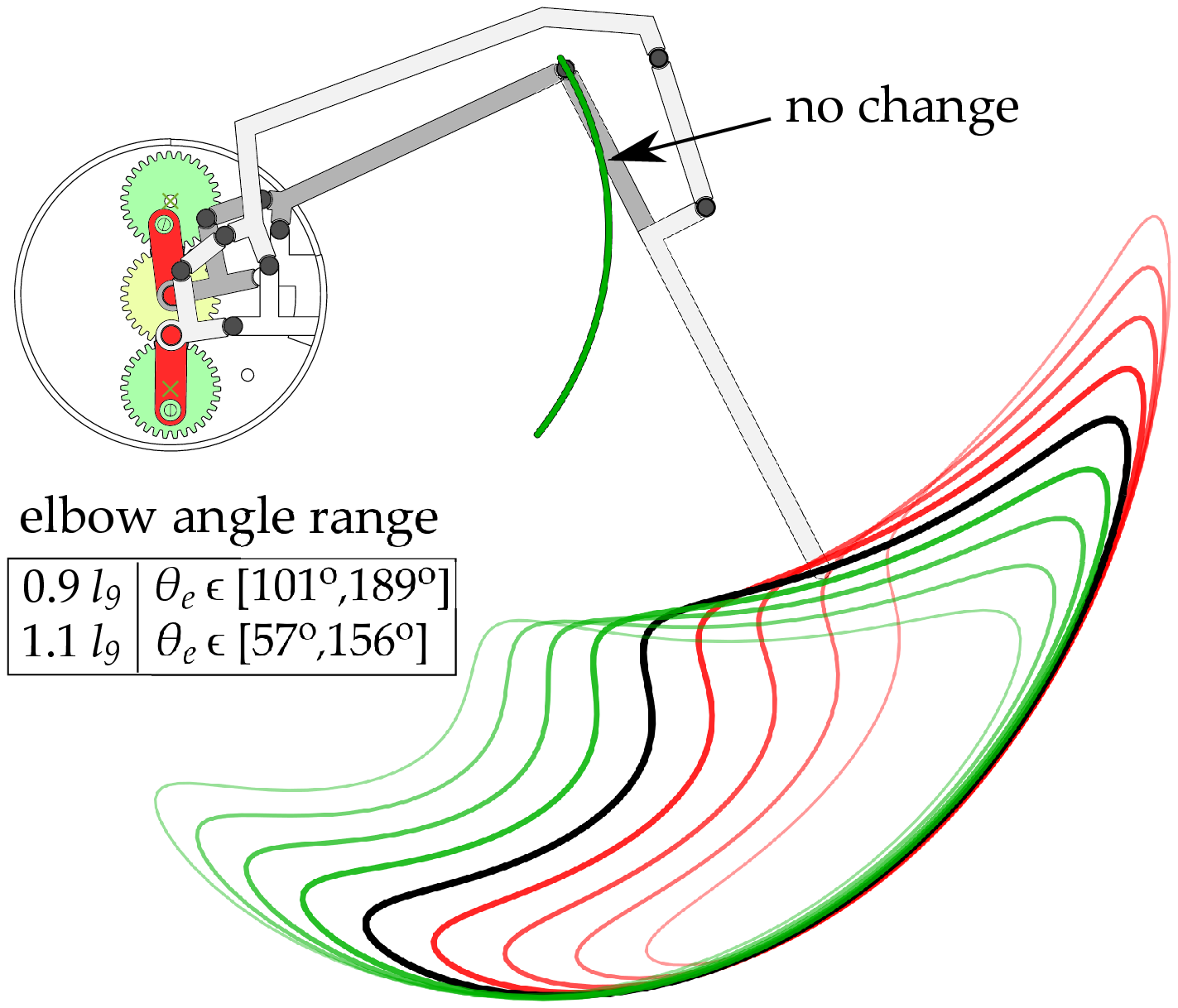}%
    }
    \caption{Armwing elbow joint and wingtip trajectories generated by varying one of the design parameter. The parameter is varied between 90\% and 110\% of its original values. The green and red trajectory colors indicate larger and smaller parameter values, respectively. Highly sensitive parameters such as case (b) and (c) lead to a huge change in the resulting flapping trajectory.}
    \label{fig:plot_sensitivity}
\end{figure*}

A considerable displacement can occur at the KS joints under high dynamic loads, therefore it is crucial to analyze the effects of the joints displacements to the resulting flapping trajectory. We evaluated this using kinematics sensitivity analyses where we evaluating how a small change in conformation parameters lead to a change in the resulting trajectory, as shown in Fig.~\ref{fig:plot_sensitivity}. The green and red lines represent an 2.5\% increase and decrease compared to the original parameter value, respectively. 
Highly sensitive parameters, such as Fig.~\ref{fig:plot_sensitivity_2} and Fig.~\ref{fig:plot_sensitivity_3}, can greatly influence the resulting flapping trajectory from a very small change in conformation.

\subsection{Structural Analysis}

We performed FEA in order to analyze the structural properties of the KS under load and simulate the material bending during KS articulation. The FEA was done using Solidworks Simulation where we subjected the input arms of the KS with some load and analyzed the resulting deformation and stress distribution within the KS. We simulated the elastic material in the KS using the material property of FLX9870 (Table \ref{tab:hinge_material}) which has the density of 1.15 g/cm$^3$ and average tensile strength of 5 MPa \cite{polyjet_datasheet}. The simulation is done with the hyperelastic Mooney-Rivlin model using the material constants of a rubber shown in \cite{shahzad2015mechanical}, with a Poisson's ratio of 0.4999, and first and second material constants of 0.3339 Mpa and -0.337 kPa, respectively.

\begin{figure}[t]
    \centering
    \subfloat[Wing expansion before the downstroke.]{%
    \label{fig:plot_fea_a}\includegraphics[clip,width=0.45\columnwidth]{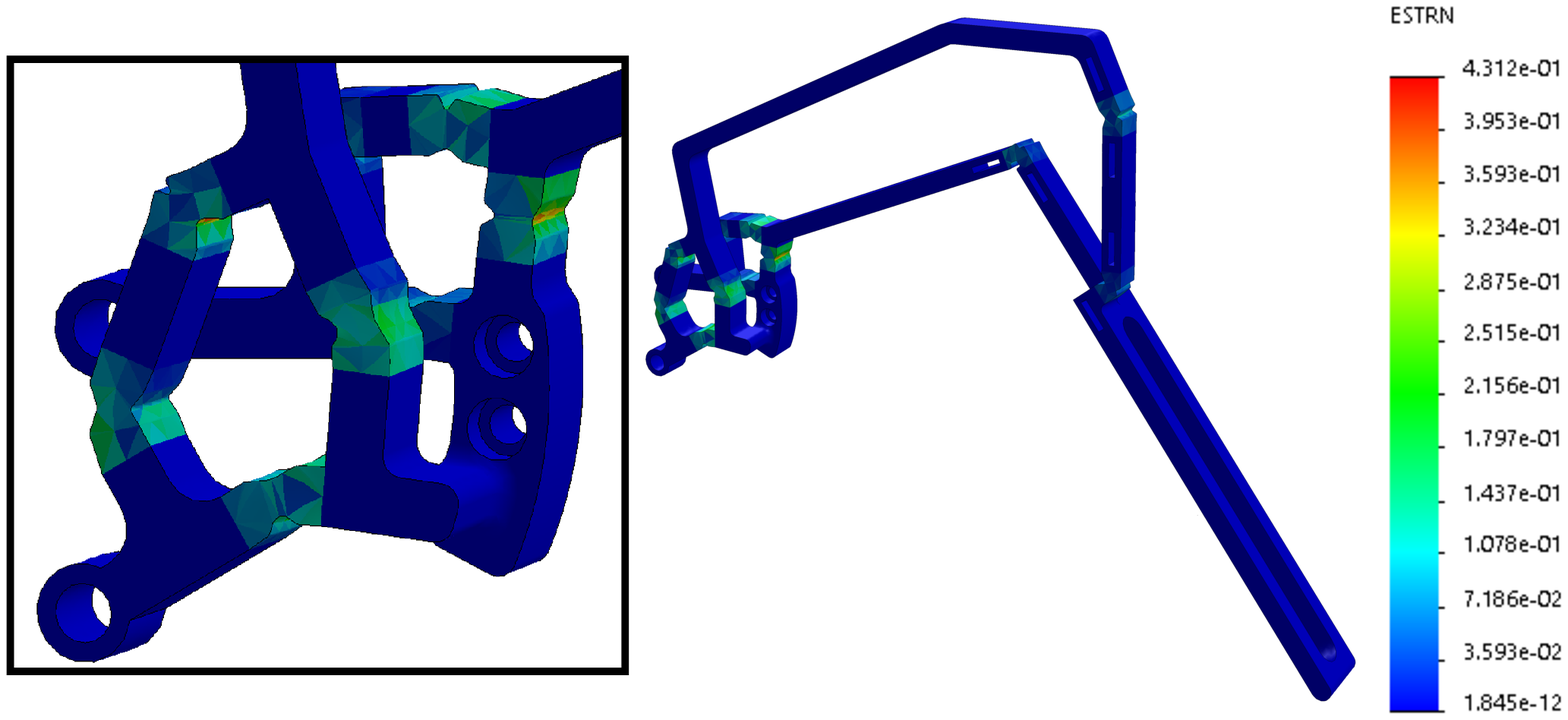}%
    }
    \quad
    \subfloat[Wing retraction before the upstroke.]{%
    \label{fig:plot_fea_b}\includegraphics[clip,width=0.45\columnwidth]{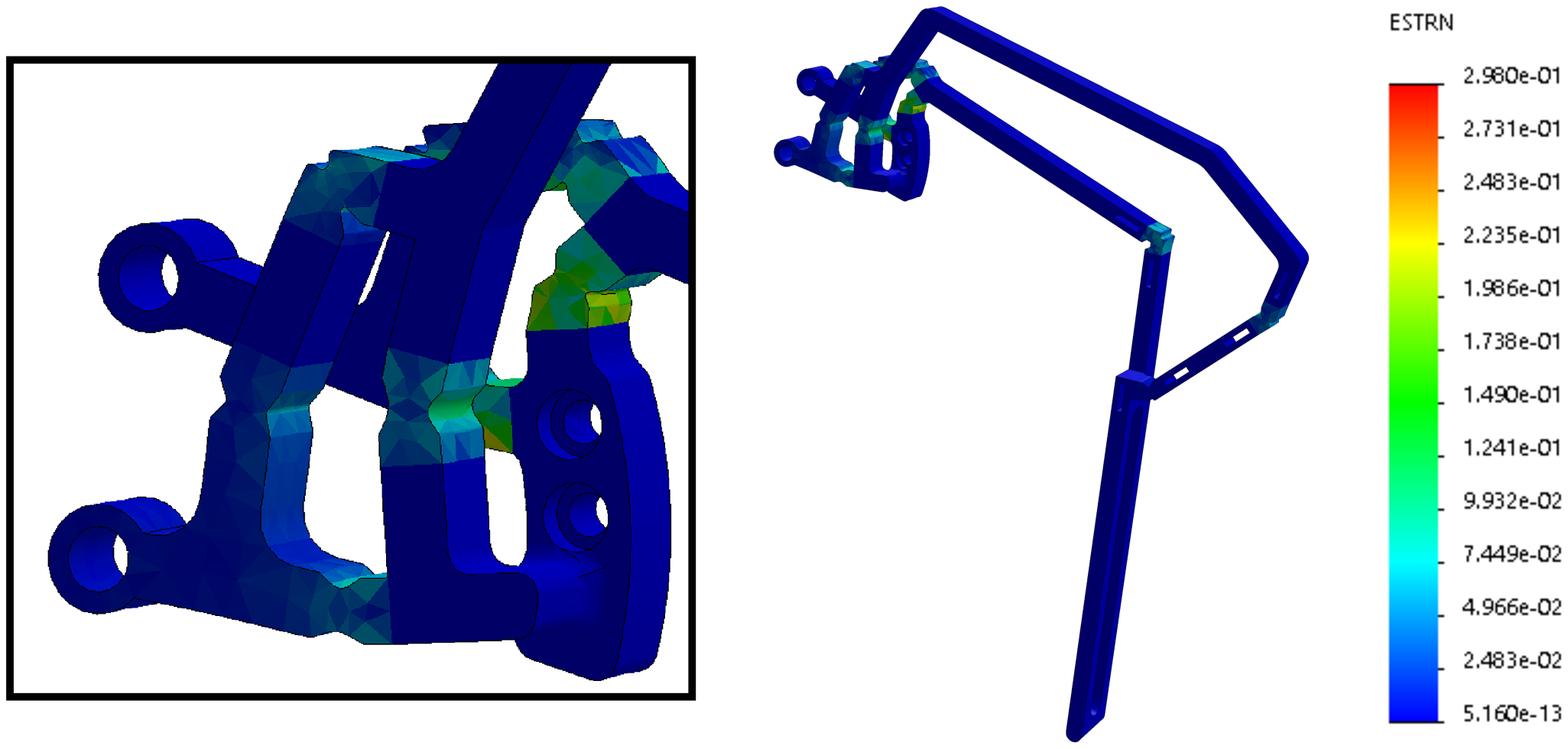}%
    }
    \vspace{0.1in}
    \caption{FEA of the compliant mechanism under static torques acting on the crank arms. The strain values are shown at the right side.}
    \label{fig:plot_fea}
\end{figure}

We simulated the deformations during the beginning of downstroke and upstroke which is timing where the KS switches direction. As shown in Fig. \ref{fig:plot_fea}, the maximum strain during these two motions are 43\% and 30\%, respectively. This is well under the maximum elongation at break of 120-140\% for this material which is sufficient to deal with unforeseen strain due to dynamic load and off-plane perturbations.

\section{CONCLUSIONS AND FUTURE WORK}

We present a novel bio-inspired monolithic bat armwing structure with both flexible and rigid materials. This armwing structure is designed to expand and retract during the wing flapping motion to maximize the net lift produced by the wing. We also presented structural and sensitivity analysis of the armwing structure to show the feasibility of using the flexible polyjet material under dynamic load and the sensitive parameters in the mechanism. These sensitive parameters can be exploited where we can incorporate \textit{morphological computation} where a simple control action can influence and achieve complex manipulation in the body morphology which is commonly seen in nature \cite{hauser_role_2012}. A small change in a sensitive conformation parameter can result a large change in flapping trajectory which affects the resulting aerodynamic forces. This can be used in a feedback stabilization controller which we will have to incorporate in the system due to the inherent pitch instability of a tail-less ornithopter such as Aerobat.



 

\bibliography{references_local, references-eric} 
\bibliographystyle{spiebib} 

\end{document}